\documentclass[runningheads]{llncs}
\usepackage[T1]{fontenc}
\usepackage[hidelinks]{hyperref}
\usepackage{graphicx}
\usepackage{amsmath,amssymb}
\usepackage{graphicx}
\usepackage{booktabs}
\usepackage{array}
\usepackage{multirow}
\usepackage{tikz}
\usetikzlibrary{positioning,fit,arrows.meta,shapes.geometric,backgrounds}

\begin{document}
\title{Federated Learning for Early Prediction of EV Charging Demand}
\titlerunning{Federated Learning for Early Prediction of EV Charging Demand}
\author{Vasilis Perifanis\inst{1}\orcidID{0000-0003-3915-9628} \and
Foteini Nikolaidou\inst{1}\orcidID{0009-0002-5691-9510} \and
Nikolaos Pavlidis\inst{1}\orcidID{0000-0001-9370-5023} \and 
Panagiotis Thomakos\inst{1}\orcidID{0009-0009-0296-8879} \and
Andreas Sendros\inst{1}\orcidID{0000-0002-6337-4320}}
\authorrunning{V. Perifanis et al.}
\institute{Indigma Innovations\\
\email{\{vasilis,f.nikolaidou,nikos,p.thomakos,andreas\}@indigma.eu}}
\maketitle              % typeset the header of the contribution
\begin{abstract}
Accurate forecasting of electric vehicle (EV) charging demand is critical for grid stability, infrastructure planning, and real-time charging optimization. In this work, we study the problem of \textbf{early prediction of charging demand}, where the total energy of a session is estimated using only information available at plug-in time and during the first minutes of charging. This enables actionable decisions while the session is still in progress, which is of direct importance for EV network operators. We construct a session-level dataset from the Adaptive Charging Network (ACN) combining session metadata with early-window charging measurements, and derive tabular features capturing user intent, temporal patterns, and initial charging behavior. We focus on a single operational depot (Caltech) and model intra-depot heterogeneity through station-level client partitions while evaluating multiple model families using a \textbf{Federated Learning} (FL) setting. Our results show that federated models can approach centralized predictive performance while keeping data in-depot, enabling privacy-enhanced training across distributed charging infrastructures. Overall, we demonstrate that reliable demand estimates can be obtained early in the session with minimal data, and that FL provides a practical pathway towards scalable and privacy-aware analytics for EV charging networks. Our code is available at: \url{https://github.com/Indigma-Innovations/federated-learning-ev-charging-demand}.

\end{abstract}
\section{Introduction}

The rapid adoption of electric vehicles (EVs) is reshaping transportation and placing new operational demands on charging infrastructures. This is evident in \emph{depot-based} environments, where vehicles must be charged concurrently under shared power constraints, limited time windows, and evolving operational priorities~\cite{al2024charging}. For fleet operators, logistics providers, and large charging-site managers, efficient charging directly affects schedules, operating costs, and service reliability. In such settings, even modest improvements in future predictions can improve charger utilization and support more effective short-term power allocation~\cite{lyu2023co}.

A key challenge is that many operational decisions must be made \emph{before} a charging session has fully elapsed. In practice, operators benefit most from \textbf{early estimates of total session energy}, obtained shortly after plug-in, while there is still time to react~\cite{akshayaa2025session}. If the expected energy demand of a session can be inferred from plug-in context and the first minutes of charging behavior, then charging policies can be adapted while the session is in progress. This creates opportunities for earlier load balancing, improved prioritization across vehicles, and tighter integration with depot-level energy management~\cite{akshay2024power,you2024fmgcn}. By contrast, predictions made late in the session may be accurate, but are far less actionable.

At the same time, deploying predictive models in charging infrastructures raises practical constraints. Charging data are naturally distributed across stations, and moving all local measurements into a centralized pipeline can be undesirable from both a systems and a governance perspective. Even within a single depot, local training remains operationally attractive, as it maintains data locality. Consequently, this motivates learning schemes that enable collaboration without requiring centralized data access.~\cite{yin2025prediction,tritsarolis2026electric}.

\textbf{Federated Learning} (FL)~\cite{mcmahan2017communication} offers a natural solution in this context. Under the FL setting, a machine learning (ML) model is trained locally at participating clients, while a central server coordinates the process by aggregating model updates. This approach allows a shared predictive model to be learned without transmitting raw data. An additional advantage of FL is its ability to be integrated directly into the charging workflow, enabling model training close to where charging sessions occur. For EV charging operators, such a setup is attractive because it supports privacy-aware learning, reduces the need for continuous raw-data transfer, and matches the inherently distributed structure of charging infrastructures.

In this paper, we study \textbf{early prediction of EV charging demand} as a supervised regression task. Specifically, given information available at connection time and during an initial observation window, the goal is to predict the total energy delivered by the session. We build a session-level dataset using the Adaptive Charging Network (ACN) framework~\cite{lee2019acn} by combining session metadata with features extracted from the \emph{first 10 minutes} of charging activity, including current, pilot signal, utilization, and trend statistics. This design reflects an operationally realistic early-decision setting, where predictions must be formed with limited information but still be accurate enough to support action~\cite{marzbani2023electric}.

We focus on one charging site (Caltech) and study \emph{intra-depot} FL, where clients are defined through station-level partitioning. This setup evaluates federated training under data distributions across multiple charging points within a single operational depot. We compare centralized training against FL across multiple model families for energy regression using features available in the early stages of charging session. ML models include linear regression (LR), the boosting-based XGBoost (XGB) model, Multi-Layer Perceptrons (MLPs), a tabular convolution neural network (CNN), a tabular Gated Recurrent Unit (GRU), and a tabular Transformer. Our goal is to evaluate predictive performance and to assess whether privacy-enhanced distributed learning can provide a practical alternative to conventional centralized training in EV charging environments. The main contributions of this work are as follows:
\begin{itemize}
    \item We formulate an \textbf{early-session regression problem} for EV charging, where the total delivered energy is predicted using only plug-in context and the first minutes of session behavior.
    \item We build a realistic session-level dataset using ACN framework and derive a compact tabular representation that captures temporal context, user-provided information, and early charging dynamics.
    \item We evaluate centralized and station-partitioned FL across multiple model families, showing that privacy-enhanced training can approach centralized performance while remaining suitable for edge-oriented deployment.
\end{itemize}

The remainder of this paper is organized as follows. Section~\ref{sec:related_work} reviews related work. Section~\ref{sec:system_model} presents the system model and learning formulation. Section~\ref{sec:experiments} describes the experimental setup and reports the results. Finally, Section~\ref{sec:conclusion} concludes the paper and outlines future directions.

\section{Related Work}
\label{sec:related_work}

Research on EV charging prediction spans both \emph{aggregate} forecasting, e.g., station or network-level charging demand and occupancy, and \emph{session-level} prediction tasks, where quantities such as delivered energy, charging duration, or flexibility are inferred for operational decision support~\cite{ostermann2024probabilistic}. Aggregate forecasting has been studied at multiple spatial and temporal scales, often including probabilistic formulations for short-term demand estimation. Session-level prediction, in contrast, focuses on learning mappings from plug-in context, historical behavior, and exogenous signals to outcomes such as total delivered energy.

Several studies have shown that session-level energy and duration can be predicted effectively under centralized training, while also highlighting the role of heterogeneity across users, stations, and charging environments~\cite{shahriar2021prediction}. Related work has also linked session-level prediction to flexibility estimation and scheduling, emphasizing that predictions must be made early enough to remain actionable~\cite{bellizio2025machine}. From a systems perspective, charging data are naturally distributed across stations and operators, which motivates privacy-aware collaborative learning. Recent work has explored FL for EV charging, mainly in time-series problems such as occupancy prediction and load forecasting across multiple sites, often with mechanisms to handle client heterogeneity~\cite{yin2025prediction,hallak2025adaptive,douaidi2025federated}. However, federated formulations have been studied less extensively for \emph{early-session} regression of total delivered energy using only a short initial observation window.

In this work, we study intra-depot FL for early-session energy prediction, where each station acts as a client and predictors are restricted to plug-in context and the first minutes of charging behavior. This complements prior centralized session-level prediction studies and extends FL-based EV charging analytics beyond occupancy and load forecasting.

\section{System Model}
\label{sec:system_model}

\subsection{Problem Definition}
\label{subsec:problem_definition}

We consider a depot-based EV charging environment in which a fleet of vehicles arrives over time and initiates charging sessions at a set of available charging stations. Each session evolves as a time-dependent process characterized by plug-in context, infrastructure metadata, and charging measurements collected during execution. From the operator's perspective, the objective is to obtain an accurate estimate of the \emph{total} energy demand of a charging session as early as possible, while the session is still in progress and corrective actions remain feasible. 

Formally, let $i$ denote a charging session with connection time $t_i^{\mathrm{conn}}$. For each session, we observe an input vector $\mathbf{x}_i \in \mathbb{R}^d$ constructed from information available at plug-in time and from measurements collected during an initial observation window of length $W$ minutes after connection. The learning objective is to estimate the total delivered energy $y_i = E_i^{\mathrm{tot}}$, where $E_i^{\mathrm{tot}}$ is the total energy delivered during the full charging session, expressed in kWh. The resulting supervised learning problem is to learn a function $f : \mathbf{x}_i \mapsto y_i$.

In this work, we focus on the \textbf{early-session prediction} setting, where only information available in the interval $[t_i^{\mathrm{conn}},\, t_i^{\mathrm{conn}} + W]$ can be used to construct predictors. This restriction prevents information leakage from later stages of the session and matches the operational setting in which charging decisions must be made early.

\begin{figure}[t]
    \centering
    \includegraphics[width=.8\columnwidth]{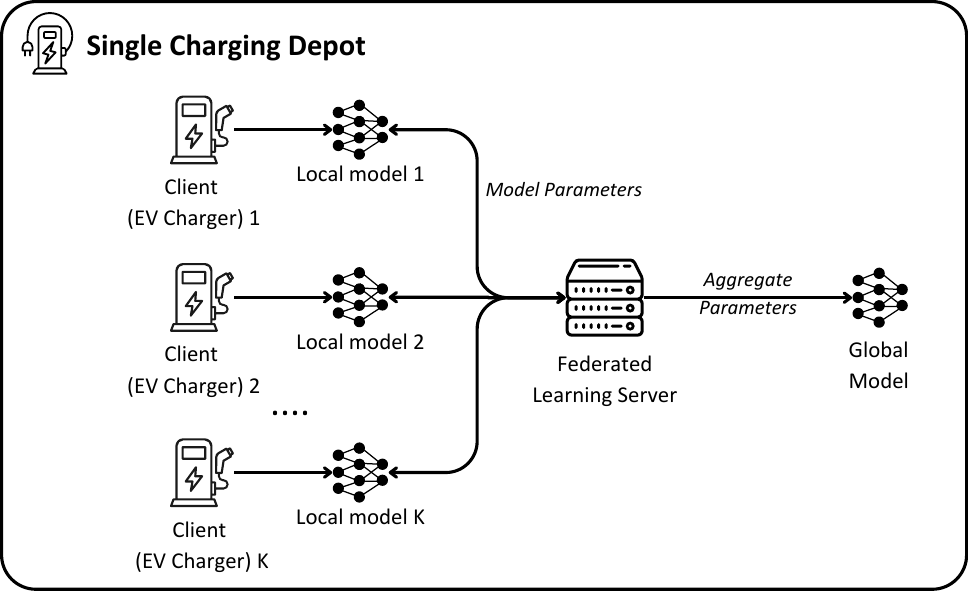}
    \caption{Station-level intra-depot FL setup. Each charging station acts as an FL client, and only parameter updates are exchanged with the FL server.}
    \label{fig:intradepot_fl_scenarios}
\end{figure}

We study the problem under two learning settings. In the first, all available data are pooled into a \emph{centralized} training pipeline. In the second, training is performed through \emph{FL} over station-level clients, where each station contributes local model updates while retaining raw data locally. The station-partitioned federated setup is illustrated in Fig.~\ref{fig:intradepot_fl_scenarios}.

\subsection{Dataset Construction}
\label{subsec:dataset_construction}

To support the early-session prediction task, we construct a session-level dataset from the ACN repository~\cite{lee2019acn}. Each sample corresponds to one charging session and combines session metadata with charging measurements, enabling prediction of total delivered energy from both plug-in context and the initial dynamics of the charging process. The dataset is constructed by combining two sources of information available for each charging session. The first source contains \emph{session-level metadata}, including session identifiers, infrastructure identifiers, timestamps, and optional user-provided inputs available at plug-in time. The second source contains \emph{time-series charging measurements}, namely timestamped observations of charging current and pilot signal recorded during the session. Only sessions for which both contextual metadata and charging measurements are available are retained. As a result, each sample in the final dataset combines static plug-in context with the early dynamic behavior of the corresponding charging process.

From the session metadata, we extract infrastructure identifiers, timestamps, and target information. Additional calendar variables are derived from the connection timestamp, namely the connection hour, weekday, month, day of year, and a binary weekend indicator. These variables capture regular temporal patterns in session demand. When available, optional user-provided fields are also incorporated, including requested energy, available charging time, and requested departure time. From these, we derive the requested departure offset relative to connection time, expressed in minutes.

Since the target task is early-session prediction, time-series features are extracted only from the initial charging period. Let $W$ denote the early-window duration. For session $i$, only signal values recorded in the interval $[t_i^{\mathrm{conn}},\, t_i^{\mathrm{conn}} + W]$
are considered for feature construction. In this work, we define $W = 10 \text{ minutes}.$ This design ensures that all predictors correspond to information that would realistically be available shortly after plug-in. It also avoids using features from later stages of the session, which would artificially simplify the learning task.

For each session, two time-series signals may be available within the early window, i.e., i) \textbf{charging current}, denoted as $I_i(t)$, and ii) \textbf{pilot signal}, denoted as $P_i(t)$. From the early charging current sequence, we compute summary statistics including the mean, maximum, standard deviation, minimum, and last observed value. The same set of statistics is computed for the pilot signal. To capture temporal evolution, we also estimate a linear trend coefficient for each signal by fitting a least-squares line over the corresponding early-window samples. These slope features summarize increasing or decreasing charging behavior during the first minutes of the session.

To quantify the interaction between observed current and available pilot capacity, we define an early-window utilization ratio at timestamps where both signals are available and the pilot signal is strictly positive $\left(u_i(t) = \frac{I_i(t)}{P_i(t)}\right)$. From this ratio, we extract the mean and maximum observed utilization. In addition, we derive an approximate early delivered energy by integrating the current signal over time under a nominal voltage assumption of $208$~V~\cite{lee2019acn}. The instantaneous power is approximated as $\widehat{P}_i(t) = \frac{208 \cdot I_i(t)}{1000}$, expressed in kW, and the corresponding early-window energy is estimated through trapezoidal integration
$\left(\widehat{E}_i^{\mathrm{early}} \approx \int_{t_i^{\mathrm{conn}}}^{t_i^{\mathrm{conn}}+W} \widehat{P}_i(t)\,dt\right)$.
This variable serves as an informative predictor summarizing the charging activity observed during the first part of the session.

Since time-series availability may vary across sessions, we include features that quantify early-window coverage. Specifically, we include the number of early current and pilot measurements, the number of merged timestamps in the early window, and the observed duration, in minutes, between the earliest and latest available timestamps in the window.
\begin{table*}[t!]
    \centering
    \caption{Summary of the feature categories used for early-session energy prediction.}
    \label{tab:feature_summary}
    \small
    \begin{tabular}{p{2.0cm} p{6cm} p{4.0cm}}
        \toprule
        \textbf{Category} & \textbf{Representative variables} & \textbf{Description} \\
        \midrule
        Identifiers \& grouping &
        Session ID, site name, station ID, station-level client ID &
        Infrastructure and client partition variables \\
        \midrule
        Plug-in context &
        Connection timestamp, connection hour, day of week, month, day of year, weekend indicator &
        Temporal context at session start \\
        \midrule
        User-provided fields &
        Requested energy (kWh), available charging time (minutes), requested departure offset from connection (minutes) &
        Optional user-side charging intent and constraints \\
        \midrule
        Early current features &
        Mean, max, min, standard deviation, first, last current, current slope per second &
        Summary and trend features from early charging current \\
        \midrule
        Early pilot features &
        Mean, max, min, standard deviation, first, last pilot, pilot slope per second &
        Summary and trend features from early pilot signal \\
        \midrule
        Interaction features &
        Mean utilization ratio, max utilization ratio &
        Ratio-based features combining current and pilot \\
        \midrule
        Early energy \& coverage &
        Approximate early-window energy (kWh), number of current samples, number of pilot samples, number of merged window samples, observed window duration (minutes) &
        Energy proxy and data-quality indicators \\
        \midrule
        Target &
        Delivered energy (kWh) &
        Total delivered session energy \\
        \bottomrule
    \end{tabular}
\end{table*}

After merging the session-level and time-series sources, we retain only sessions that satisfy the following conditions: (i) the session exists in both sources, (ii) the target variable is available, and (iii) the number of early current measurements is at least five. The resulting dataset is therefore a session-level cohort, suitable for both centralized and federated model training. The final feature space contains contextual, behavioral, interaction, and grouping variables. Table~\ref{tab:feature_summary} provides the summary of the main feature categories used in this work.

\subsection{Intra-Depot Federated Partitioning and Data Heterogeneity Analysis}
\begin{figure*}[t]
    \centering
    \includegraphics[width=0.7\columnwidth]{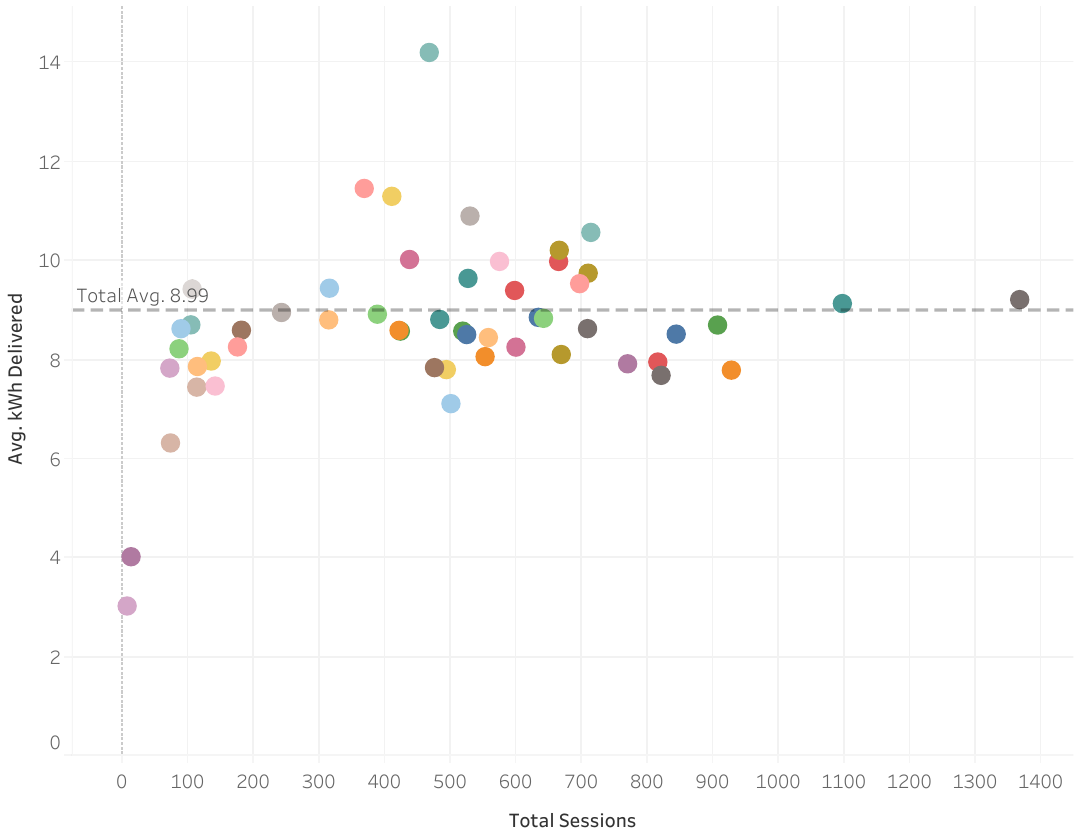}
    \caption{Station-level relationship between total number of sessions and average delivered energy. The dashed line indicates the global depot average ($\approx 9$~kWh).}
    \label{fig:scatter_plot}
\end{figure*}
\label{subsec:intra_depot_partitioning}
In this section, we define the intra-depot federated partitioning scheme and analyze the resulting statistical heterogeneity across clients. Specifically, we model each charging station as an individual federated client and examine how data distributions vary within a single operational depot. Each station contributes its local subset of sessions and participates in collaborative optimization through FedAvg~\cite{mcmahan2017communication} while retaining raw data locally. This setup preserves data locality and reflects a realistic deployment scenario in which measurements are generated and stored at the station level. For the specific depot (Caltech) we focus on, this results in $K=54$ clients with varying sample sizes.

\begin{figure*}[t]
    \centering
    \includegraphics[width=0.7\columnwidth]{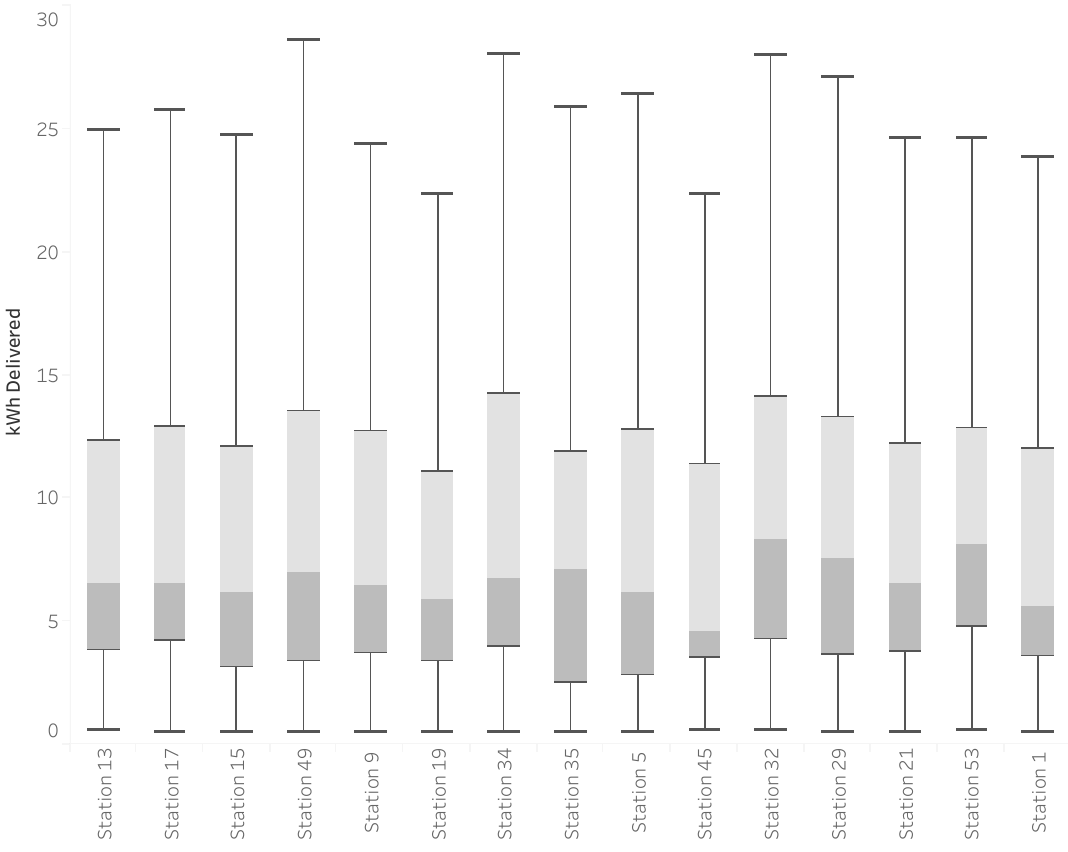}
    \caption{Distribution of delivered energy across the top-15 station-level clients.}
    \label{fig:boxplot}
\end{figure*}

We first examine the relation between station activity and delivered energy. Fig.~\ref{fig:scatter_plot} presents the average delivered energy per station against the total number of sessions. Most stations cluster around the global average of approximately $9$~kWh, indicating broadly consistent behavior across the depot, although a small number of stations exhibit systematically higher or lower average demand. Fig.~\ref{fig:boxplot} further shows the distribution of delivered energy across the top-15 most active stations. Although median values are broadly similar, the interquartile ranges and upper tails vary noticeably across clients. Some stations exhibit wider distributions and larger maxima, suggesting localized differences in usage patterns and session intensity.

%Taken together, the scatter and boxplot analyses indicate that the intra-depot partition is not strictly IID. The bulk of stations exhibit comparable average behavior, but there exist localized deviations in both mean energy and distributional spread. 

To quantify the statistical heterogeneity across station-level clients, we compare each client's target distribution against the global depot-level target distribution using the Jensen--Shannon (JS) divergence. Let $P_k$ denote the empirical distribution of delivered energy for client $k$, and $P$ the global distribution. The divergence is defined as:
\begin{equation}
\mathrm{JS}(P_k \,\|\, P) = \frac{1}{2} \mathrm{KL}(P_k \,\|\, M_k) + \frac{1}{2} \mathrm{KL}(P \,\|\, M_k),
\end{equation}
where $M_k = \frac{1}{2}(P_k + P)$ and $\mathrm{KL}(\cdot \,\|\, \cdot)$ denotes the Kullback--Leibler divergence. In practice, both $P_k$ and $P$ are approximated via histogram-based density estimates. To summarize heterogeneity across clients, we report both the maximum client-wise divergence $\max_k \mathrm{JS}(P_k \,\|\, P)$ and a sample-size weighted average:
\begin{equation}
\mathrm{JS}_{\text{weighted}} = \sum_{k=1}^{K} w_k \, \mathrm{JS}(P_k \,\|\, P), \quad \text{with} \quad w_k = \frac{n_k}{\sum_{j=1}^{K} n_j},
\end{equation}
where $n_k$ is the number of samples for client $k$.

To obtain a dataset-specific reference for heterogeneity, we construct a null IID baseline via a permutation-based simulation. Specifically, we repeatedly randomize the assignment of samples to clients while preserving the original client sizes $\{n_k\}_{k=1}^{K}$, recompute $\mathrm{JS}_{\text{weighted}}$ for each randomized realization, and estimate the distribution of divergence values expected under an IID assumption. Denoting the mean and standard deviation of this null distribution by $\mu_{\text{IID}}$ and $\sigma_{\text{IID}}$, respectively, we define the heterogeneity threshold as: $\tau_{\text{IID}} = \mu_{\text{IID}} + 2\sigma_{\text{IID}}$. For the Caltech depot, this procedure yields $\tau_{\text{IID}} = 0.0069$.

\begin{figure*}[t]
    \centering
    \includegraphics[width=\columnwidth]{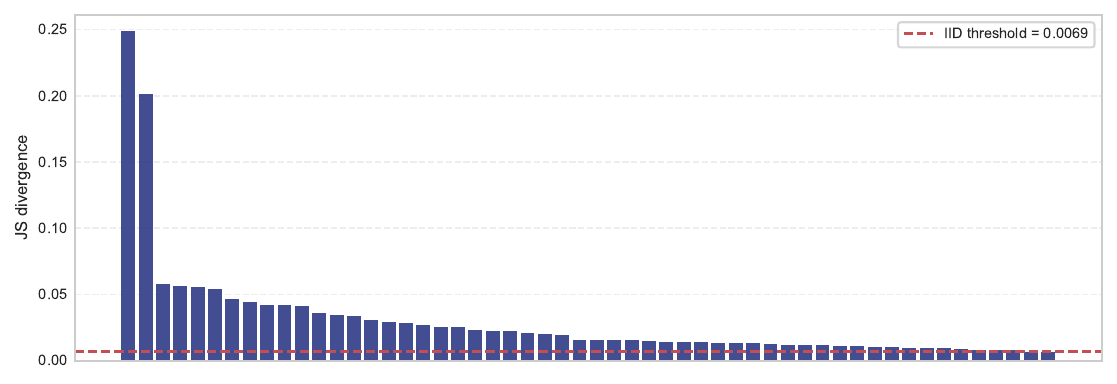}
    \caption{Ranked client-wise Jensen--Shannon divergence values against the global depot-level target distribution.}
    \label{fig:js_client_divergence}
\end{figure*}

The observed weighted JS divergence is $\mathrm{JS}_{\text{weighted}} = 0.0169$, which exceeds the dataset-specific threshold, and the partition is therefore classified as non-IID. At the same time, the ranked client-wise divergences in Fig.~\ref{fig:js_client_divergence} indicate that the degree of heterogeneity is generally moderate. Most stations exhibit relatively small divergence values, while only a small number of clients emerge as clear outliers, with the maximum observed divergence reaching $0.2486$. Furthermore, the most divergent clients tend to correspond to stations with comparatively small sample counts, suggesting that part of the observed heterogeneity is driven by low-data clients rather than broad systematic shifts across the entire depot. Overall, the intra-depot station partition is characterized as mildly non-IID, exhibiting measurable but not severe distributional heterogeneity.

\section{Experimental Setup}
\label{sec:experiments}

\subsection{Learning Settings}
\label{subsec:learning_settings}

We evaluate the early-session regression task under two learning settings. In the \textbf{centralized} setting, all training samples are pooled and a single model is learned from the complete training set. In the \textbf{federated} setting, training is performed with FedAvg over station-level clients defined by the intra-depot partitioning of Section~\ref{subsec:intra_depot_partitioning}. Thus, each client corresponds to an individual station and participates in collaborative training without sharing raw session data.

All experiments are conducted on the session-level dataset described in Section~\ref{subsec:dataset_construction}. We use a random $70/15/15$ split into training, validation, and test subsets. Numerical features are standardized using statistics from the training split, while temporal variables derived from the connection timestamp are encoded through cyclical sine/cosine transformations. Categorical identifiers are represented through embedding layers in the neural models.

We compare two dummy baselines, two ML regressors, and four neural network architectures for tabular data. The same model definitions and hyperparameter settings are used in both centralized and federated training. Table~\ref{tab:compared_models} summarizes the models considered in this work. All neural models are optimized with Adam using a learning rate of $10^{-3}$ and a batch size $128$. In centralized training, models are trained for $40$ epochs. In the federated setting, we use FedAvg for $400$ communication rounds, with $3$ local epochs per selected client and a client sampling fraction of $20\%$ per round. In both settings, the checkpoint with the best validation score is retained for final evaluation. Results are reported as mean $\pm$ standard deviation over $10$ runs with different random seeds.
\begin{table*}[t]
\centering
\small
\caption{Models evaluated and configurations used.}
\label{tab:compared_models}
\begin{tabular}{p{1.2cm}p{2.0cm}p{8.8cm}}
\toprule
\textbf{Family} & \textbf{Model} & \textbf{Configuration (default)} \\
\midrule
Dummy
& Mean-only predictor 
& Predicts the mean of the training targets for all samples. \\
Dummy
& Gaussian predictor 
& Samples predictions from $\mathcal{N}(\mu_{\text{train}}, \sigma_{\text{train}})$ estimated from the training targets. \\
\midrule
ML
& LR
& Simple linear regression baseline. \\
ML
& XGB
& $400$ estimators, learning rate $0.05$, max depth $6$, subsample $0.9$, colsample\_bytree $0.9$. \\
\midrule
Neural
& MLP
& Tabular MLP with learned categorical embeddings and a 4-layer feed-forward head: input $\rightarrow 128 \rightarrow 128 \rightarrow 64 \rightarrow 1$, ReLU activations, dropout $0.2$, Softplus output. \\
Neural
& CNN
& Token-based 1D CNN with numerical feature projections and categorical embeddings of dimension $16$, two Conv1D layers with $32$ hidden channels, global average pooling, dropout $0.1$, and Softplus output. \\
Neural
& GRU 
& Token-based GRU with numerical feature projections and categorical embeddings of dimension $16$, single GRU layer with hidden dimension $32$, regression head $32 \rightarrow 32 \rightarrow 1$, dropout $0.1$, and Softplus output. \\
Neural
& Transformer 
& Token-based Transformer with numerical and categorical token embeddings of dimension $64$, $4$ attention heads, $2$ encoder layers, learned \texttt{[CLS]} token, regression head $64 \rightarrow 64 \rightarrow 1$, dropout $0.1$, and Softplus output. \\
\bottomrule
\end{tabular}
\end{table*}

Predictive performance is evaluated on the test set using \textbf{Mean Absolute Error (MAE)} and \textbf{Root Mean Squared Error (RMSE)}. MAE provides an interpretable absolute error in delivered energy units, while RMSE penalizes larger deviations more strongly.

\subsection{Results}
We organize the evaluation around two questions: (i) how accurately can total session energy be predicted from the first $10$ minutes of charging, and (ii) how closely can the station-partitioned FL approach the centralized reference.

\begin{figure}[t]
    \centering
    \includegraphics[width=\textwidth]{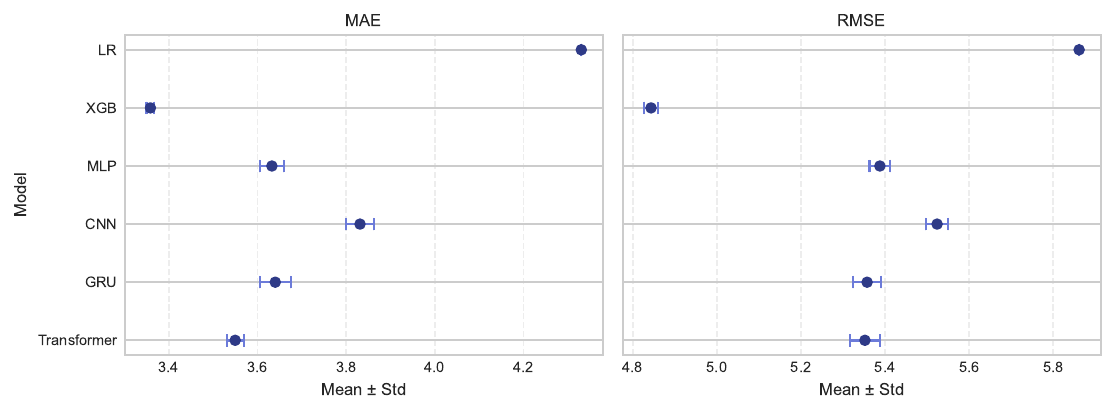}
    \caption{Centralized test set summary for MAE and RMSE across evaluated models (mean and standard deviation over $10$ runs).}
    \label{fig:centralized_test_metrics}
\end{figure}

\paragraph{Centralized Performance}
Fig.~\ref{fig:centralized_test_metrics} summarizes centralized test performance. XGB achieves the strongest results, with MAE $3.36 \pm 0.01$ and RMSE $4.84 \pm 0.02$. Among the neural models, Transformer performs best (MAE $3.55 \pm 0.02$, RMSE $5.35 \pm 0.04$), followed by the MLP, GRU, and CNN. LR is the weakest learned model (MAE $4.33$, RMSE $5.86$), indicating that the relation between early-session descriptors and final delivered energy is not captured by a linear model. All learned models outperform the dummy baselines by a wide margin. The mean-only dummy reaches MAE $5.711$ and RMSE $8.035$, while the Gaussian dummy performs worse (MAE $7.901$, RMSE $10.650$). These results confirm that the proposed early-session representation contains substantial predictive signal and that useful demand estimation is feasible from the first minutes of charging.
\begin{figure}[t]
    \centering
    \includegraphics[width=\linewidth]{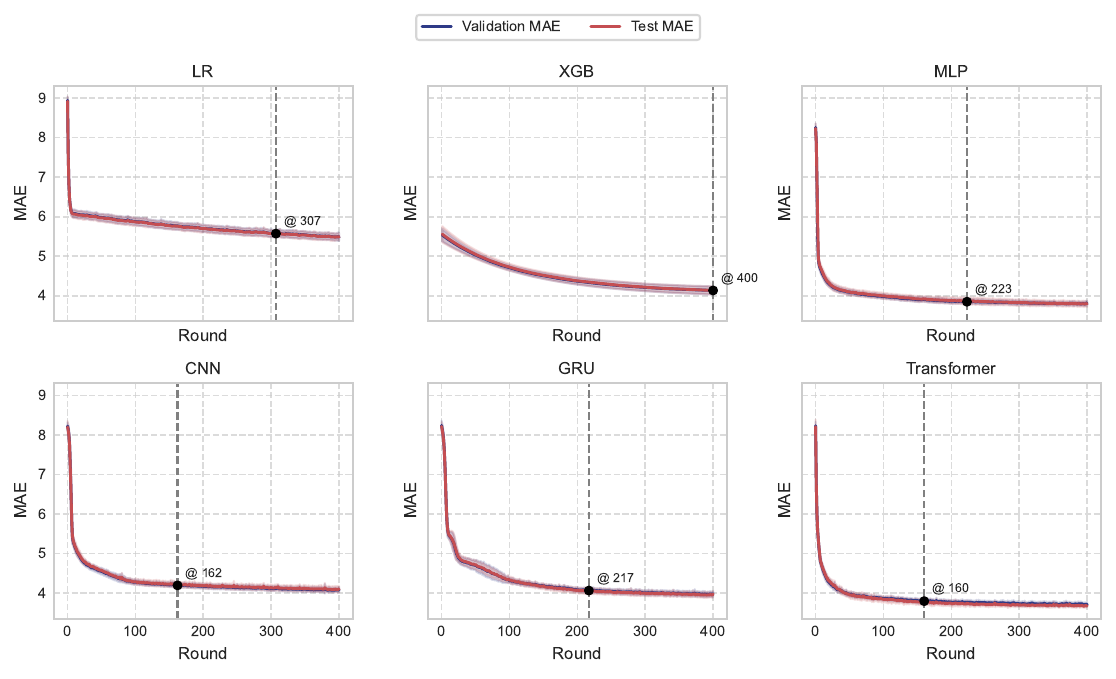}
    \caption{Federated training convergence curves. Solid lines indicate the mean MAE across seeds, while shaded regions denote the standard deviation. The vertical dashed line denotes the estimated convergence round.}
    \label{fig:federated_convergence}
\end{figure}

\paragraph{Federated Learning Performance}
We next compare station-partitioned FL against the centralized reference. Fig.~\ref{fig:federated_convergence} shows the global validation and test MAE across communication rounds for all federated models. 

First, all federated models exhibit a steep reduction in error during the early communication rounds, indicating that collaborative training quickly captures a substantial fraction of the predictive signal. The neural models converge faster than the ML baselines, with the Transformer and CNN stabilizing earliest, followed closely by the GRU and MLP. In contrast, LR and XGB improve more gradually and require more rounds before reaching their best operating point. 

Second, the validation and test curves remain closely aligned across models, suggesting that the federated optimization process is stable and does not exhibit overfitting across rounds. The relatively narrow shaded regions for most models further indicate modest variation across random seeds. 

Third, the convergence rounds differ across architectures. The Transformer converges at round $160$, the CNN at round $162$, the GRU at round $217$, the MLP at round $223$, and LR at round $307$, whereas XGB had not yet converged within the $400$ rounds considered. This is important from a deployment perspective, since communication efficiency is a central concern in federated systems.

Table~\ref{tab:federated_vs_centralized} summarizes the final predictive performance under both training settings. Among the federated models, Transformer achieves the best accuracy (MAE $3.69 \pm 0.05$, RMSE $5.56 \pm 0.14$), followed by the MLP (MAE $3.79 \pm 0.08$), the GRU (MAE $3.99 \pm 0.01$), and the CNN (MAE $4.14 \pm 0.08$). In contrast, federated LR and federated XGB remain weaker, with large performance gaps relative to their centralized counterparts. This suggests that the neural models are better able to absorb the heterogeneity induced by station-level data partitioning. A second important result is that the federated degradation relative to centralized training is moderate for the top performing neural models. The Transformer increases from $3.55$ to $3.69$ in MAE, and the MLP from $3.63$ to $3.79$. These gaps are small enough to support the practical viability of FL. By comparison, the degradation is more evident for XGB and LR, indicating that not all model families transfer equally well to the federated regime.

In addition to predictive performance, Table~\ref{tab:federated_vs_centralized} reports peak RAM usage and serialized model size. These quantities are crucial in federated deployments, where on-device training may take place on resource-constrained edge hardware~\cite{perifanis2023towards}. The results show a clear trade-off between accuracy and model efficiency. Simpler models, such as LR, have a negligible storage footprint but substantially weaker predictive performance. On the other hand, the Transformer achieves the strongest accuracy, but it is also the largest model in terms of on-disk size and exhibits the highest memory usage among the evaluated neural architectures. MLP and GRU provide a more balanced operating point, combining competitive federated accuracy with substantially smaller model footprints. 

Overall, the federated results support three conclusions. First, collaborative learning across stations can preserve much of the predictive performance achieved under centralized training, even though raw data remain decentralized. Second, model choice matters, e.g., while XGB is the strongest centralized baseline, transformer- and MLP-based models are more effective under station-partitioned federated optimization. Third, the reported memory and storage footprints indicate that the FL paradigm can be supported on edge-oriented infrastructures, with lightweight models offering an attractive compromise between predictive quality, privacy and deployability.
\begin{table*}[t]
\centering
\caption{Comparison of federated and centralized performance across models. Resource metrics include peak RAM usage during training and serialized model size.}
\label{tab:federated_vs_centralized}
\resizebox{\textwidth}{!}{
\begin{tabular}{lcccccc}
\toprule
\textbf{Model} 
& \multicolumn{2}{c}{\textbf{Federated}} 
& \multicolumn{2}{c}{\textbf{Centralized}} 
& \textbf{RAM (GB)} 
& \textbf{Model Size (MB)} \\
\cmidrule(lr){2-3} \cmidrule(lr){4-5}
& \textbf{MAE} & \textbf{RMSE} 
& \textbf{MAE} & \textbf{RMSE} 
& & \\
\midrule
LR          & 5.53 $\pm$ 0.09 & 8.10 $\pm$ 0.20 & 4.33 $\pm$ 0.00 & 5.86 $\pm$ 0.00 & 0.78 & 0.01 \\
XGB         & 4.13 $\pm$ 0.11 & 6.17 $\pm$ 0.19 & 3.36 $\pm$ 0.01 & 4.84 $\pm$ 0.02 & 1.10 & 1.50 \\
MLP         & 3.79 $\pm$ 0.08 & 6.41 $\pm$ 2.09 & 3.63 $\pm$ 0.03 & 5.39 $\pm$ 0.03 & 1.23 & 0.14 \\
CNN         & 4.14 $\pm$ 0.08 & 6.56 $\pm$ 2.08 & 3.83 $\pm$ 0.03 & 5.52 $\pm$ 0.03 & 1.25 & 0.05 \\
GRU         & 3.99 $\pm$ 0.01 & 5.71 $\pm$ 0.15 & 3.64 $\pm$ 0.04 & 5.36 $\pm$ 0.03 & 1.25 & 0.05 \\
Transformer & 3.69 $\pm$ 0.05 & 5.56 $\pm$ 0.14 & 3.55 $\pm$ 0.02 & 5.35 $\pm$ 0.04 & 1.64 & 2.33 \\
\bottomrule
\end{tabular}
}
\end{table*}

\section{Conclusion}
\label{sec:conclusion}
In this paper, we studied early-session prediction of EV charging demand using information available at plug-in time and during the first $10$ minutes of charging. The results show that accurate early prediction is feasible using a tabular representation of initial charging behavior. Under centralized training, XGBoost achieved the strongest performance, while under federated training the Transformer and MLP were the most competitive neural models. These findings suggest that federated learning can serve as a practical and privacy-aware alternative for EV charging analytics in distributed infrastructures. Future work will extend the present intra-depot setting to cross-depot federated learning across multiple charging sites. Additional directions include the study of alternative aggregation methods beyond FedAvg, the evaluation of other model families, richer time-series modeling of early charging trajectories, and tighter integration with depot-level optimization. Specifically, combining early charging-demand prediction with complementary signals such as photovoltaic production, local storage, or site-level load forecasts may enable more informed and energy-aware decision support for EV depot operation.

\begin{credits}
\subsubsection{\ackname} This work was partially supported by the project ``Open CloudEdgeIoT Platform Uptake in Large Scale Cross-Domain Pilots (O-CEI)'' funded by the European Commission under Project code/Grant Number 101189589 through the HORIZON program.

\subsubsection{\discintname} The authors have no competing interests to declare that are relevant to the content of this article.
\end{credits}

%\begin{credits}
% \subsubsection{\ackname} A bold run-in heading in small font size at the end of the paper is
% used for general acknowledgments, for example: This study was funded

% \subsubsection{\discintname}
% It is now necessary to declare any competing interests or to specifically
% state that the authors have no competing interests. Please place the
% statement with a bold run-in heading in small font size beneath the
% (optional) acknowledgments\footnote{If EquinOCS, our proceedings submission
% system, is used, then the disclaimer can be provided directly in the system.},
% for example: The authors have no competing interests to declare that are
% relevant to the content of this article. Or: Author A has received research
% grants from Company W. Author B has received a speaker honorarium from
% Company X and owns stock in Company Y. Author C is a member of committee Z.
% \end{credits}
\bibliographystyle{splncs04}
\bibliography{references}
\end{document}